\documentclass[]{spie}  

\usepackage{amsmath,amsfonts,amssymb}
\usepackage{graphicx}
\usepackage[colorlinks=true, allcolors=blue]{hyperref}
\usepackage{todonotes}

\title{Ultrasound Image Generation using Latent Diffusion Models}

\author[1]{Benoit Freiche}
\author[1]{Anthony El-Khoury}
\author[1]{Ali Nasiri-Sarvi}
\author[1]{Mahdi S. Hosseini}
\author[2]{Damien Garcia}
\author[2]{Adrian Basarab}
\author[3]{Mathieu Boily}
\author[1]{Hassan Rivaz}
\affil[1]{Concordia University, Montreal, Canada}
\affil[2]{CREATIS, Villeurbanne, France}
\affil[3]{McGill University, Montreal, Canada}

\authorinfo{Further author information: (Send correspondence to Benoit Freiche)\\B.F.: E-mail: benoit.freiche@concordia.ca,\\}

\begin{document} 
\maketitle

\begin{abstract}
Diffusion models for image generation have been a subject of increasing interest due to their ability to generate diverse, high-quality images. Image generation has immense potential in medical imaging because open-source medical images are difficult to obtain compared to natural images, especially for rare conditions. The generated images can be used later to train classification and segmentation models. In this paper, we propose simulating realistic ultrasound (US) images by successive fine-tuning of large diffusion models on different publicly available databases. To do so, we fine-tuned Stable Diffusion \cite{rombach:CVPR:2022}, a state-of-the-art latent diffusion model, on BUSI (Breast US Images) \cite{alDhabyani:DIB:2020} an ultrasound breast image dataset. We successfully generated high-quality US images of the breast using simple prompts that specify the organ and pathology, which appeared realistic to three experienced US scientists and a US radiologist. Additionally, we provided user control by conditioning the model with segmentations through ControlNet \cite{zhang:ICCV:2023}. We will release the source code at \url{http://code.sonography.ai/} to allow fast US image generation to the scientific community.
\end{abstract}


\section{INTRODUCTION}

Image generation is important in the medical imaging with deep learning (DL) community, as access to a substantial amount of labeled images is complex and costly. Increasing DL models with automatically generated images has been shown to enhance overall training performance \cite{khosravi:EBM:2024,sagers:arxiv:2023}. Today, ultrasound (US) image generation is predominantly based on physics-based models, such as MUST \cite{MUST_Garcia}, Field II \cite{Field_Jensen},  k-waves \cite{treeby2010k} and UltraRay \cite{duelmer:arxiv:2025}. 
While these models are effective for image generation \cite{behboodi2019ultrasound}, they are computationally expensive. In addition, they struggle to model pathologies or tailor images to a specific organ. Various models have been explored for US image generation \cite{dominguez:MICCAI:2024,stojanovski:IWASMU:2023,Zhang2023DiffusionMD,chou:arxiv:2025,reynaud:Miccai:2024}, however, they often lack generalization capabilities and image quality.  

\begin{figure}[h!]
\centering
\includegraphics[width=0.7\columnwidth]{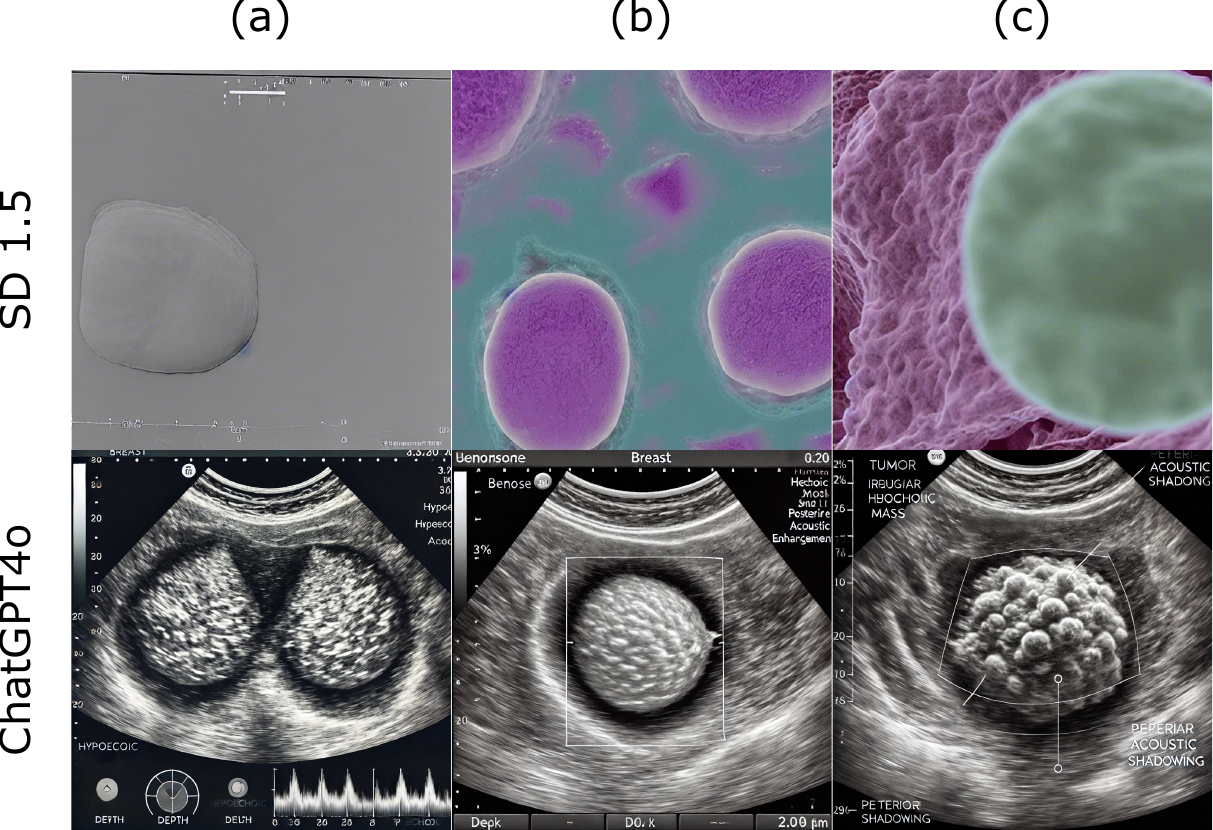}
\caption{Current results of two models for image generation: (first row) Stable Diffusion 1.5 (second row) ChatGPT4o. Different prompts have been used: (a) ``Ultrasound image of breast'', (b) ``Ultrasound image of breast with a benign lesion'', (c) ``Ultrasound image of breast with a malignant tumor''} 
\label{stable_diffusion_image_heart}
\end{figure}

The computer vision community has recently introduced numerous diffusion models for high-quality image generation. Despite these advancements, these models are not yet able to generate realistic US images, as illustrated in Figure \ref{stable_diffusion_image_heart}. Promising approaches for US image generation have recently been developed using diffusion models, either with physics-inspired models \cite{dominguez:MICCAI:2024} or by guiding the image generation with segmentations \cite{stojanovski:IWASMU:2023}. These are small models learned from scratch, lacking the representation power of larger ``foundation'' models released by the computer vision community. The application of diffusion models to improve the quality of ultrasound images is also an active field of research\cite{zhang:TUFFC:2024,asgariandehkordi:TUFFC:2024,stevens:TMI:2024,asgariandehkordi:IUS:2023}.

It would be valuable to explore the application of large foundation models, which have been trained on extensive GPU resources and massive databases, to generate high-quality US images. 

In this paper, we propose exploring the feasibility of US image generation from a text prompt by fine-tuning a state-of-the-art latent diffusion model on BUSI (Breast cancer US images) \cite{alDhabyani:DIB:2020}. We also propose to condition the images using segmentation masks with ControlNet.\cite{zhang:ICCV:2023}.

\section{METHODS}
\subsection{Stable Diffusion}

Diffusion models, popularized by Denoising Diffusion Probabilistic Models\cite{ho:NEURIPS:2020} (DDPM), have redefined the gold standard in generative modeling, achieving unparalleled performance and stability compared to traditional approaches such as GANs \cite{dhariwal:NEURIPS:2021}. They learn how to iteratively denoise data starting from Gaussian noise to generate high-quality samples. Text-to-image diffusion models translate textual descriptions into high-quality, semantically accurate images. In this paper, we fine-tuned Stable Diffusion \cite{rombach:CVPR:2022} on two datasets to generate realistic US images. Stable Diffusion is a latent diffusion model, which means that the diffusion process operates in the latent space of an autoencoder, and it uses CLIP \cite{radford:ICML:2021} for text conditioning. The projection to the latent space is made through a variational autoencoder, which compresses the 512*512 pixels input images to 64*64 pixels latent images on which the diffusion process will occur. The denoising process is learned using a U-NET \cite{ronneberger:MICCAI:2015}. Learning the diffusion process in the latent space leads to easier training, requiring less computational resources than usual models with comparable performances. For instance, we could fine-tune the model on our datasets in a few hours using a single A100 GPU.

\subsection{ControlNet}\label{ControlNET}

\begin{figure}[h!]
\centering
\includegraphics[width=0.7\columnwidth]{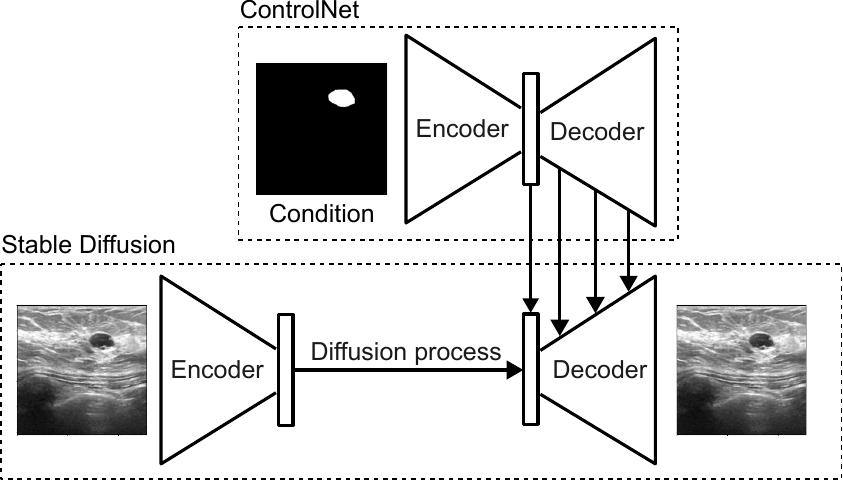}
\caption{Demonstration of the main principles of Stable Diffusion and ControlNet. The ControlNet is a copy of the autoencoder of Stable Diffusion, on which zero-convolution layers are added and progressively learned.} 
\label{ControlNet}
\end{figure}

The input of original diffusion models, such as Stable Diffusion, is a text prompt specifying the characteristics of the image. In  2023, Zhang et al. introduced ControlNet \cite{zhang:ICCV:2023}, a model that builds on a pre-trained diffusion model (e.g., Stable Diffusion) to specify conditional inputs (such as sketches or edge maps in the original paper) to constrain the generated output precisely. In order to add better user control over the output, our contribution is to use the segmentation masks available in our databases as the condition. This allows the user to generate several images corresponding to a specific segmentation. ControlNet works in the following way: the weights of the original diffusion model are frozen, and a copy of the network is trained with the condition as input. This network copy output is then added to the output of each decoding layer. Initially, the links are done through zero-convolution layers, which allows a smooth integration of the condition to the original generative model. Figure \ref{ControlNet} shows a scheme of this process. 

\subsection{Dataset}
We used a publicly available dataset to fine-tune Stable Diffusion: BUSI \cite{alDhabyani:DIB:2020}, which contains 780 US images, divided into three classes: normal (133 images), benign (437 images), and malignant (210 images) cases. This dataset also provides the corresponding segmentations for the benign and malignant cases.

\section{RESULTS AND DISCUSSION}
\subsection{Results of Stable Diffusion on BUSI}
\subsubsection{Qualitative results}
\begin{figure}[h!]
\centering
\includegraphics[width=\columnwidth]{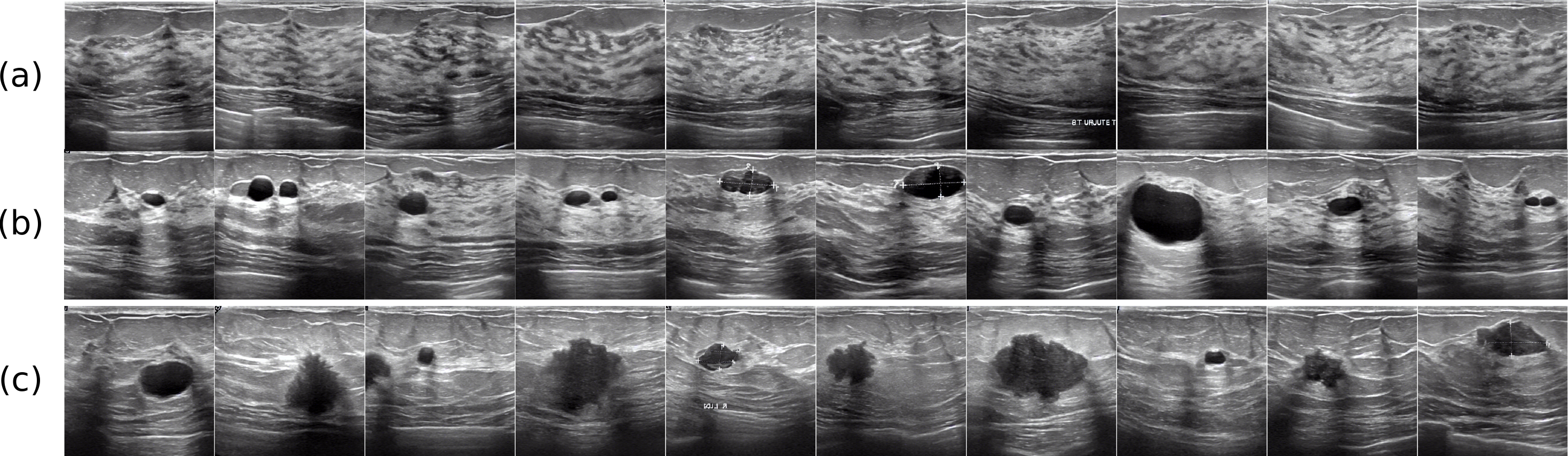}
\caption{Some results of our fine-tuning on BUSI images. Each row represents a category, (a) normal, (b) benign, (c) malignant. The results are realistic to US experts and an experienced US radiologist. The categories (Normal, Benign, and Malignant) are also respected.} 
\label{BUSI_SD}
\end{figure}

At first, we fine-tuned Stable Diffusion on BUSI. BUSI is a rather simple dataset, as the variability in the images is relatively low. A few samples are displayed in Figure \ref{BUSI_SD}. We used the prompt ``Ultrasound image of a normal/benign/malignant breast'' for the training. Our images are close to the training images, and we can distinguish the different structures: skin, a layer of fat and glandular tissue, pectoral muscle, and ribs, in some cases. The image texture is close to what can be found in US images. The backgrounds are similar between the different categories. The malignant and benign tumors are distinguishable, the benign ones being round-shaped black cysts, while the malignant ones have irregular shapes. One can also see that some of our images contain white lines of measures and writings, similar to the ones sometimes present in regular US images of the breast.

\subsubsection{Quantitative results}

Classification networks were trained using different combinations of BUSI and BUSI-generated images to assess the quality and diversity of the generated samples. Here are some notable results obtained with Resnet-50 (training setup outlined in \cite{nasiri:MICCAI:2024}): (1) Training with a dataset augmented by some of the generated images led to an improvement from 81\% to 87\% AUC while testing on B\cite{yap:JBHI:2017} (a different Breast Imaging dataset). This result shows that training models with generated images can improve the classification results. (2) A Resnet-50 was trained with generated images from a Stable Diffusion model trained using only 20\% of the BUSI dataset. It achieved 94\% AUC while predicting the generative model's training set, showing that Stable Diffusion had encapsulated the inner statistics of breast cancer US images. 

\subsubsection{Adding condition with segmentations}

\begin{figure}[h!]
\centering
\includegraphics[width=\columnwidth]{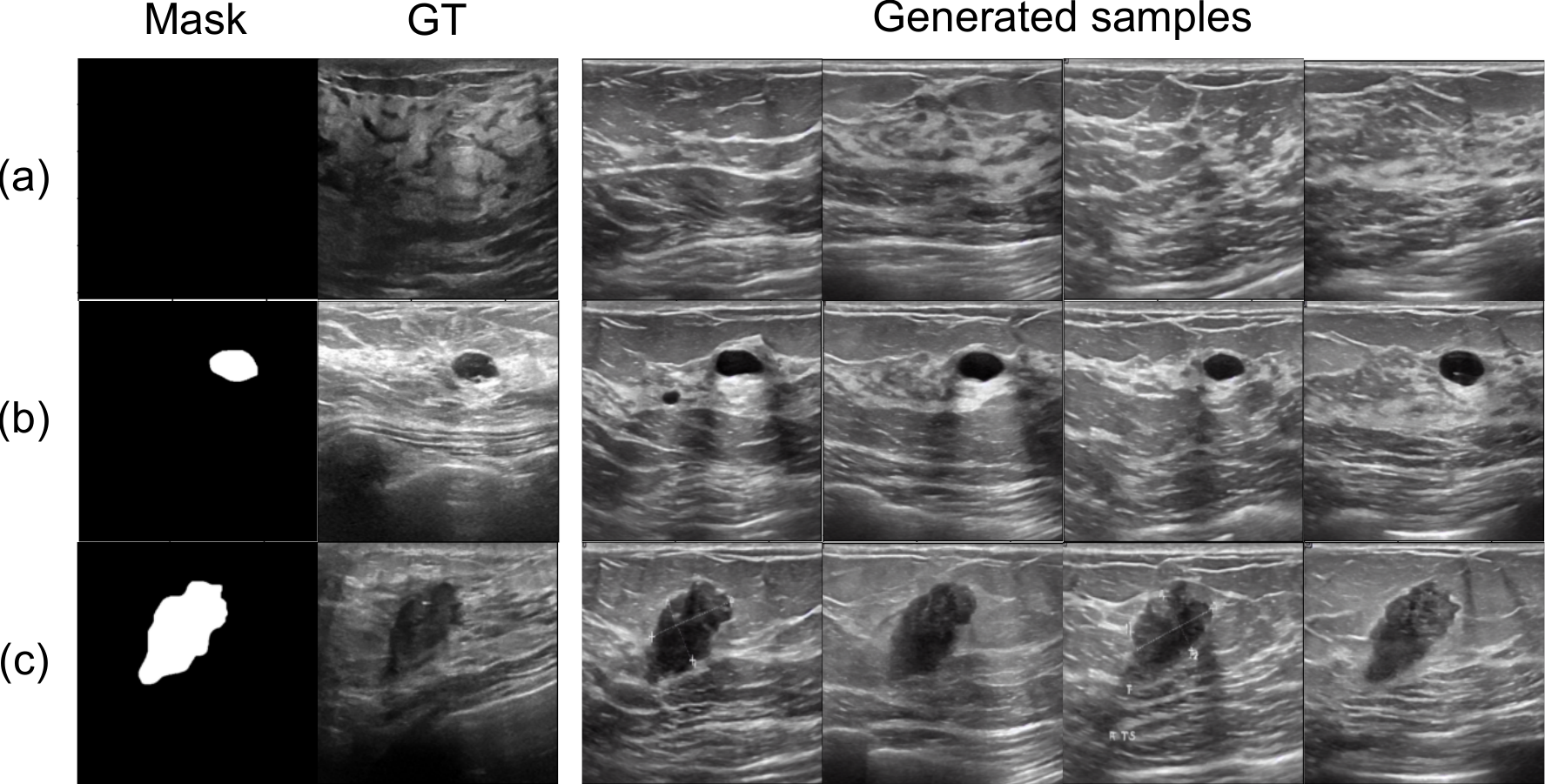}
\caption{Samples conditioned by segmentation masks, generated with ControlNet on BUSI. Each row represents a category: normal (a), benign (b), malignant (c). The columns are from left to right: (1) the input segmentation masks (2) the original image corresponding to the segmentation (ground truth) (3-6) four generated samples. For the benign and malignant cases, the lesion corresponds to the input segmentation mask.} 
\label{BUSI_ControlNet}
\end{figure}

As described in Section \ref{ControlNET}, we also used ControlNet to condition the output with segmentation. Similar ideas for US image generation have been discussed in \cite{stojanovski:IWASMU:2023,dominguez:MICCAI:2024}. The ControlNet was trained with the generative network already fine-tuned on BUSI. Some examples of the results we obtained are displayed in Figure \ref{BUSI_ControlNet}. The Figure displays the mask we used for the samples' generation, the ground truth image from the dataset, and the generated samples. It can be seen that the generated samples correspond to the imposed segmentation mask, showing that ControlNet was able to impose a condition for the creation of new plausible US images.

\subsection{Discussion and Conclusions}

This study represents an initial exploration into the feasibility of developing a latent diffusion model capable of generating realistic and diverse US images. By fine-tuning the Stable Diffusion v1.5 model, we generated high-quality breast cancer images. Furthermore, we used segmentation masks to condition the image generation process, allowing the user to generate specific patterns. 

Despite these advancements, there remains considerable scope for further improvement. Firstly, utilizing large models such as Stable Diffusion presents significant memory challenges during fine-tuning. These challenges may be mitigated by implementing more efficient fine-tuning techniques \cite{jia:ECCV:2022} or better optimizers.  

In terms of data, enhancing the model's generalization capabilities could be achieved by training it on a diverse array of US image databases, thus constructing a multi-task network proficient in comprehending the distinctive features of US images. 

Furthermore, creating a dataset comprising physics-based simulated images generated via tools such as MUST or Field II presents a promising avenue for exploration. Such a dataset would enable the network to establish correlations between physical conditions and the resultant images, facilitating the rapid generation of US images corresponding to specific physical conditions and diffusion distributions.

Moreover, increasing user control over the image generation process would further enhance the model's utility. Two main paths can be followed here: First, allow for the specification of desired modifications for an already generated sample. Second, a more profound work on the conditioning part can also be considered. 

Ultimately, we aim to make the trained model available to the broader research community for simulating US images. Such model can be used to generate databases for ultrasound image segmentation \cite{goudarzi:TBE:2023},  registration \cite{zhou:JBHI:2016} and cancer detection \cite{harmanani:arxiv:2025}. Recent work on combining the power of diffusion models with active ultrasound sensing \cite{van:TUFFC:2024} further amplifies the utility of these models.

\acknowledgments 
 
This project is funded by the Government of Canada’s New Frontiers in Research Fund (NFRF), [NFRFE-2022-00295], and the Natural Sciences and Engineering Research Council of Canada (NSERC).

\bibliography{report} 
\bibliographystyle{spiebib} 

\end{document}